\documentclass[10pt,twocolumn,letterpaper]{article}

\usepackage{cvpr}
\usepackage{times}
\usepackage{epsfig}
\usepackage{graphicx}
\usepackage{amsmath}
\usepackage{amssymb}
\usepackage{multirow}
\usepackage{subfigure}
\usepackage{algorithm}
\usepackage{algorithmic}
\usepackage[pagebackref=true,breaklinks=true,letterpaper=true,colorlinks,bookmarks=false]{hyperref}

\cvprfinalcopy % *** Uncomment this line for the final submission

\def\cvprPaperID{1608} % *** Enter the CVPR Paper ID here

\ifcvprfinal\fi
\begin{document}

%%%%%%%%% TITLE
\title{\ Large Margin Object Tracking with Circulant Feature Maps}

\author{Mengmeng Wang$^{1}$, Yong Liu$^{1}$, Zeyi Huang$^{2}$\\
$^{1}$ Institute of Cyber-Systems and Control, Zhejiang University, $^{2}$ Exacloud Limited, Zhejiang, China\\
{\tt\small mengmengwang@zju.edu.cn; yongliu@iipc.zju.edu.cn; haoran@qunhemail.com}
}

\maketitle
%\thispagestyle{empty}

%%%%%%%%% ABSTRACT
\begin{abstract}
 Structured output support vector machine (SVM) based tracking algorithms have shown favorable performance recently. Nonetheless, the time-consuming candidate sampling and complex optimization limit their real-time applications. In this paper, we propose a novel large margin object tracking method which absorbs the strong discriminative ability from structured output SVM and speeds up by the correlation filter algorithm significantly. Secondly, a multimodal target detection technique is proposed to improve the target localization precision and prevent model drift introduced by similar objects or background noise. Thirdly, we exploit the feedback from high-confidence tracking results to avoid the model corruption problem. We implement two versions of the proposed tracker with the representations from both conventional hand-crafted and deep convolution neural networks (CNNs) based features to validate the strong compatibility of the algorithm. The experimental results demonstrate that the proposed tracker performs superiorly against several state-of-the-art algorithms on the challenging benchmark sequences while runs at speed in excess of 80 frames per second.
\end{abstract}

%%%%%%%%% BODY TEXT
\section{Introduction}

Visual tracking enjoys a wide popularity recently and has been applied in many applications such as robotic services, surveillance, human motion analyses, human-computer interactions and so on. In this paper, we consider the most general scenario of visual tracking, i.e., short-term, single-object tracking with the target given in the first frame. The most difficult point of this problem is to track the target at a high speed for real-time applications while handle all challenging factors simultaneously both from background or the target itself such as occlusions, deformations, fast motions, illumination variations and so on.

Due to the lack of training samples, most existing trackers handle this problem from two aspects. The first one is to explore an effective tracking algorithm which can be designed to be either discriminative \cite{kalal2012tracking,danelljan2014adaptive,hare2011struck,icml2015_hong15,ningobject,babenko2011robust,zhang2014meem} or generative \cite{ross2008incremental,kwon2010visual,ASLA2012visual,SCM2014robust} models. It seeks to design a robust classifier or filter to detect the target, and establish an optimal mechanism to update the model at each frame. The other one is to exploit the power of the target representation which may come from conventional handcraft features \cite{ASLA2012visual,kwon2010visual,kalal2012tracking,danelljan2014adaptive,ross2008incremental} or high-level convolutional features \cite{icml2015_hong15,STCT2016stct,HDT2016hedged,FCNT2015visual,HCF2015hierarchical} from deep Convolutional Neural Networks (CNNs). These methods improve performance significantly from different aspects. However, to further improve the performance by more complex tracking algorithms or features, it would undoubtedly increase the computational complexity, which would limit the real-time performance of visual tracking.

The most popular and successful framework for visual tracking is tracking-by-detection \cite{henriques2015high,kalal2012tracking,zuo2016learning,hare2011struck,ningobject,hare2011struck} which treats the tracking problem as a detection task and learns information about the target from each detection online. There are many classification algorithms used in this framework, such as multiple instance learning \cite{babenko2011robust}, P-N learning \cite{kalal2012tracking}, online boosting \cite{grabner2006real,grabner2008semi}, support vector machines (SVM) \cite{hare2011struck,icml2015_hong15,ningobject,zhang2014meem} and so on. Among them, structured output SVM is demonstrated with an excellent potential in this field \cite{hare2011struck,ningobject}. Structured output SVM is a kind of classification algorithm which can deal with complex outputs like trees, sequences, or sets rather than class labels \cite{structured_SVM}. Hare et al. \cite{hare2011struck} employ this algorithm in the visual tracking for the first time and improve tracking accuracy considerably in several benchmarks \cite{wu2013online,wu2015object}. They propose a tracking algorithm named Struck based on kernelized structured output SVM where the output space is defined as the translations of the target relative to the previous frame. However, Struck suffers from a high computational complexity by its complex optimization while its training samples are still not dense enough. Therefore it operates slowly and limits to extend to higher dimensional features. Ning et al. \cite{ningobject} propose a dual linear structured SVM (DLSSVM) algorithm which approximates nonlinear kernels with explicit feature maps. DLSSVM improves tracking performance significantly, while its tracking speed is not fast enough for realtime applications, especially when scale estimation is considered, as well as feature dimensions and budgets of support vectors are increased. Thus, it is significant to design a novel tracking algorithm based on structured SVM which can not only absorbs the strong discrimination from structured SVM, but also processes sufficiently fast with higher dimensional features and more dense samples.

Recently, a group of correlation filter (CF) based trackers \cite{danelljan2014adaptive,henriques2015high,danelljan2014accurate,staple_2016_CVPR,SCT2014fast,bolme2010visual} have attracted extensive attentions due to their significant computational efficiency. CF enables training and detection with densely-sampled examples and high dimensional features in real time by using the fast Fourier transform (FFT). Since Bolme et al. \cite{bolme2010visual} introduce the CF into the visual tracking field, several extensions have been proposed to improve tracking performance. Henriques et al. \cite{henriques2015high} propose a high speed tracker with kernelized correlation filters (KCF) and multi-channel features which enables further extension for high dimensional features while remaining the real-time capability. Danelljan et al. \cite{danelljan2014accurate} figure out the fast scale estimation problem by learning discriminative CF based on a scale pyramid representation. One deficiency of CF is the unwanted boundary effects introduced by the periodic assumption for all circular shifts, that would degrade the discriminative ability of tracking models. To resolve this issue, Danelljan et al. \cite{SRDCF2015learning} introduce a spatially regularized component in the learning to penalize
CF coefficients depending on their spatial locations and achieve excellent tracking accuracy. However, this algorithm reduces the computational efficiency of CF and runs at a reported speed of 5 frames per second (FPS). The evolution of these methods motivate us to improve the discriminative ability of CF based tracking algorithm and remain its high operating speed.

With the great power in the feature representations, CNNs have been demonstrated significant success on many computer vision tasks, including visual tracking. Recent studies \cite{FCNT2015visual,HCF2015hierarchical,STCT2016stct,HDT2016hedged,icml2015_hong15} have shown state-of-the-art results on many object tracking benchmarks. Ma et al. \cite{HCF2015hierarchical} exploit features extracted from pretrained deep CNNs and learn adaptive CFs on several CNN layers to improve tracking accuracy and robustness. Wang et al. \cite{STCT2016stct} present a sequential training method for CNN that is regarded as an ensemble with each channel of the output feature map as an individual base learner. These methods validate the strong capacity of CNNs for the target representation at the cost of time consumption and high requirements of computational resources.

In this paper, we consider the problems mentioned above and propose a large margin object tracking method with circulant feature maps (LMCF). The main contributions of our work can be summarized as follows:
 \begin{itemize}
   \item We propose a novel structured SVM based tracking method which takes dense circular samples into account in both training and detection processes. A bridge was built up to link our problem with CF, which speeds up the optimization process significantly.
   \item We explore a multimodal target detection technique to prevent the model drift problem introduced by similar objects or background noise.
   \item We establish a model update strategy to avoid model corruption by the high-confidence selection from tracking results.
  % \item We implement two versions of the proposed tracker with the representations from both conventional hand-crafted and deep CNNs based features respectively. The experimental results show favorable performances against state-of-the-art methods.
 \end{itemize}

%
%\begin{figure}[t]
%\begin{center}
%   %\includegraphics[width=0.8\linewidth]{egfigure.eps}
%\end{center}
%   \caption{Example of caption.  It is set in Roman so that mathematics
%   (always set in Roman: $B \sin A = A \sin B$) may be included without an
%   ugly clash.}
%\label{fig:long}
%\label{fig:onecol}
%\end{figure}

%------------------------------------------------------------------------
\section{Large Margin Object Tracking with Circulant Feature Maps}
In this section, we first present the problem formulation of the large margin tracking method with circulant feature maps. Next, we deduce a fast optimization algorithm that builds up a bridge between our problem formulation and the well-known correlation filter. Thirdly, a multimodal target detection method is proposed to improve the localization precision and prevent model drift introduced by similar objects or background noise. In the end, we present a model update strategy by exploiting the feedback from tracking results to avoid the model corruption.
%-------------------------------------------------------------------------
\subsection{Problem formulation}

We consider the tracking-by-detection framework in this paper. When receiving a new frame, our goal is to learn a classifier which can distinguish the target from its surrounding background in real time. The employed classifier is a structured output SVM which is different from conventional binary discriminative classifiers. It can directly estimate the relative movement between adjacent frames rather than discriminate whether it is the target or not. Additionally, the structured output SVM used here is distinct from the methods \cite{hare2011struck, ningobject} in both the variable definitions and the objective function.

The object of large margin learning over structured output spaces is to learn a function $f:X \to Y$ based on the input-output pairs, where $X$ is the input spaces and $Y$ is arbitrary discrete output spaces. In our case, all the cyclic shifts of the image patch centered around the target are considered as the training samples, i.e., $Y = \left\{ {\left. {\left( {w,h} \right)} \right|w \in \left\{ {0,...,W - 1} \right\},h \in \left\{ {0,...,H - 1} \right\}} \right\}$, where $W$ and $H$ are the width and the height of the image patch. Hence, the input-output pairs are defined as $\left( {{\bf{x}},{{\bf{y}}_{w,h}}} \right)$, where ${\bf{x}} \in X$ denotes the image patch which contains and is proportional to the target bounding box at center, ${{\bf{y}}_{w,h}} \in Y$ represents its corresponding cyclic transform. With different cyclic shifts ${\bf{y}}_{w,h}$, the pairs stand for different image regions which contain diverse translated targets. The joint feature maps of these cyclic image patches are denoted as $\Psi \left( {{\bf{x}},{{\bf{y}}_{w,h}}} \right)$, whose specific form depends on the nature of the problem.

We aim to measure the compatibility between the input-output pairs $\left( {{\bf{x}},{{\bf{y}}}} \right)$ with $F:X \times Y \to \mathbb{{R}}$ from which we can acquire a prediction by maximizing $F$ over the response variable for a specific given input ${\bf{x}}$. Then the general form of the function $f$ can be denoted as
\begin{equation}\label{function_f}
f\left( {{\bf{x}};{\bf{w}}} \right) = \mathop {\arg \max }\limits_{{\bf{y}} \in Y} F\left( {{\bf{x}},{\bf{y}};{\bf{w}}} \right)
\end{equation}
where we assume $F$ to be a linear function, $F\left( {{\bf{x}},{\bf{y}};{\bf{w}}} \right) = \left\langle {{\bf{w}},\Psi \left( {{\bf{x}},{\bf{y}}} \right)} \right\rangle $ and ${\bf{w}}$ denotes the parameter vector which can be learned from the soft-margin support vector machine learning over structured outputs. $F$ can also be extended to nonlinear situation which will be discussed in the next section. We penalize margin violations by a quadratic term, leading to the following optimization problem:
\begin{equation}\label{object_function}
\begin{aligned}
&\mathop {\min }\limits_{\bf{w}} \frac{1}{2}{\left\| {\bf{w}} \right\|^2} + C\sum\limits_{w = 1}^{W - 1} {\sum\limits_{h = 1}^{H - 1} {\xi _{w,h}^2} } \\
&{\text{s}}{\text{.t}}{\text{.}}\forall w,\forall h,\forall {{\bf{y}}_{w,h}} \in Y\backslash {{\bf{y}}_{0,0}}: \\
&F\left( {{{\bf{x}}},{{\bf{y}}_{0,0}};{\bf{w}}} \right) - F\left( {{{\bf{x}}},{{\bf{y}}_{w,h}};{\bf{w}}} \right) \geqslant \sqrt {\Delta \left( {{{\bf{y}}_{0,0}},{{\bf{y}}_{w,h}}} \right)}  - {\xi _{w,h}}
\end{aligned}
\end{equation}
where ${\bf{y}}_{0,0}$ denotes the observed output with no cyclic transform and ${\xi _{w,h}}$ is the slack variable which penalizes the margin violations. The regularization parameter $C >0$ controls the trade-off between training error minimization and margin maximization. ${\Delta \left( {{{\bf{y}}_{0,0}},{{\bf{y}}_{w,h}}} \right)}$ quantifies the loss associated with a prediction ${{\bf{y}}_{w,h}}$ when the true output value is ${{\bf{y}}_{0,0}}$. We define the loss function as
\begin{equation}\label{loss_function}
\Delta \left( {{{\bf{y}}_{0,0}},{{\bf{y}}_{w,h}}} \right) = m\left( {{{\bf{y}}_{0,0}}} \right) - m\left( {{{\bf{y}}_{w,h}}} \right)
\end{equation}
where $m\left(  \bullet  \right)$ is designed to follow a Gaussian function that takes a maximum value for the centered target and smoothly reduces to 0 for larger shifts.

The optimization problem in Eq.\ref{object_function} pursues to ensure that the value of $F\left( {{\bf{x}},{{\bf{y}}_{0,0}};{\bf{w}}} \right)$ is greater than $F\left( {{\bf{x}},{{\bf{y}}_{w,h}};{\bf{w}}} \right)$, by a margin which depends on the loss function as Eq.\ref{loss_function}.

%-------------------------------------------------------------------------
\subsection{Fast online optimization}
The conventional structured SVM in visual tracking is solved by sequential minimal optimization (SMO) step \cite{hare2011struck} or the basic dual coordinate descent (DCD) optimization process \cite{ningobject}. Thus the tracking speed is limited due to their high computational complexity. Inspired by \cite{henriques2015high}, we propose a novel algorithm to employ Fourier transform to speed up the optimization.

Following the constraint in Eq.\ref*{object_function}, we reformulate it by adding Eq.\ref{added_constraint} into the constraint,
\begin{equation}\label{added_constraint}
\begin{aligned}
F\left( {{\bf{x}},{{\bf{y}}_{0,0}};{\bf{w}}} \right) - F\left( {{\bf{x}},{{\bf{y}}_{0,0}};{\bf{w}}} \right) \geqslant \sqrt {\Delta \left( {{{\bf{y}}_{0,0}},{{\bf{y}}_{0,0}}} \right)}  - {\xi _{0,0}}
\end{aligned}
\end{equation}
where ${\xi _{0,0}}$ denotes the slack variable of the true output which is set to 0. Then the optimization problem can be rewritten as
\begin{equation}\label{optimization}
\begin{aligned}
&\mathop {\min }\limits_{\bf{w}} \frac{1}{2}{\left\| {\bf{w}} \right\|^2} + C\sum\limits_{w = 0}^{W - 1} {\sum\limits_{h = 0}^{H - 1} {\xi _{w,h}^2} } \\
&{\text{s}}{\text{.t}}{\text{.}}\forall w,\forall h,\forall {{\bf{y}}_{w,h}} \in Y:\\
&F\left( {{\bf{x}},{{\bf{y}}_{0,0}};{\bf{w}}} \right) - F\left( {{\bf{x}},{{\bf{y}}_{w,h}};{\bf{w}}} \right) \geqslant \sqrt {\Delta \left( {{{\bf{y}}_{0,0}},{{\bf{y}}_{w,h}}} \right)}  - {\xi _{w,h}}
\end{aligned}
\end{equation}

For clarity, we first formulate our optimization method for the joint feature maps defined in the one-dimensional domain, i.e., set $W$ or $H$ to 1. Here we set $H=1$ and omit $h$ in the subscript temporarily. It can be generalized to two dimensions in the same way. Now Eq.\ref{optimization} is reformulated as
\begin{equation}\label{optimization_short}
\begin{aligned}
&\mathop {\min }\limits_{\bf{w}} \frac{1}{2}{\left\| {\bf{w}} \right\|^2} + C\left\| {{\bf{\zeta }}} \right\|_2^2\\
&{\text{s}}{\text{.t}}{\text{.}}\forall w,\forall {{\bf{y}}_w} \in Y:{{\bf{w}}^T}{\Phi _0} - {{\bf{w}}^T}\Phi  \geqslant \Upsilon  - {\bf{\zeta }}
\end{aligned}
\end{equation}
 where ${\bf{\zeta }} = \left[ {{\xi _0},...,{\xi _{W - 1}}} \right]$ represents the vector of slack variables. $\Phi  = \left[ {\Psi \left( {{\bf{x}},{{\bf{y}}_0}} \right),...,\Psi \left( {{\bf{x}},{{\bf{y}}_{W - 1}}} \right)} \right]$ is a circulant matrix formed by the joint feature maps of all the cyclic training samples and ${\Phi _0} = \left[ {\Psi \left( {{\bf{x}},{{\bf{y}}_0}} \right),...,\Psi \left( {{\bf{x}},{{\bf{y}}_0}} \right)} \right]$ is constructed with ${\Psi \left( {{{\bf{x}}},{{\bf{y}}_0}} \right)}$ in $W$ columns. $\Upsilon  = \left[ {\sqrt {\Delta \left( {{{\bf{y}}_0},{{\bf{y}}_0}} \right)} ,...,\sqrt {\Delta \left( {{{\bf{y}}_0},{{\bf{y}}_W-1}} \right)} } \right]$ denotes the loss vector .

 %$\mathcal F$ ${\mathcal F}^{{\text{ - }}1}$
 To solve the problem online, we define a new variable ${\bf{z}} = {\bf{\zeta }} + {{\bf{w}}^T}{\Phi _0} - {{\bf{w}}^T}\Phi  - \Upsilon $, ${\bf{z}} \geqslant 0$. Plug $\bf{z}$ into the Eq.\ref{optimization_short}:
 \begin{equation}\label{optimization_final}
 \begin{aligned}
&\mathop {\min }\limits_{\bf{w}} \frac{1}{2}{\left\| {\bf{w}} \right\|^2} + C\left\| {{{\bf{w}}^T}\Phi  - \left( {{{\bf{w}}^T}{\Phi _0} - \Upsilon  - {\bf{z}}} \right)} \right\|_2^2\\
&{\text{s}}{\text{.t}}{\text{. }}{\bf{z}} \geqslant 0
\end{aligned}
\end{equation}
 with the circulant nature of $\Phi $, we have
 \begin{equation}\label{DFT}
 {{\bf{w}}^T}\Phi  = {\left( {{{\mathcal F}^{ - 1}}\left( {{{\hat \Psi }^*}\left( {{\bf{x}},{{\bf{y}}_0}} \right) \circ {\bf{\hat w}}} \right)} \right)^T}
\end{equation}
 where ${\hat  \bullet }$ and ${\mathcal F}^{{\text{ - }}1}$ denotes the discrete Fourier transform (DFT) and its inverse, $ \circ $ represents the element-wise multiplication, ${{{\hat \Psi }^*}}$ means the complex conjugate of ${\hat \Psi }$.

\begin{figure*}[t]
\begin{center}
{\includegraphics[height=1.8in,width=7in,angle=0]{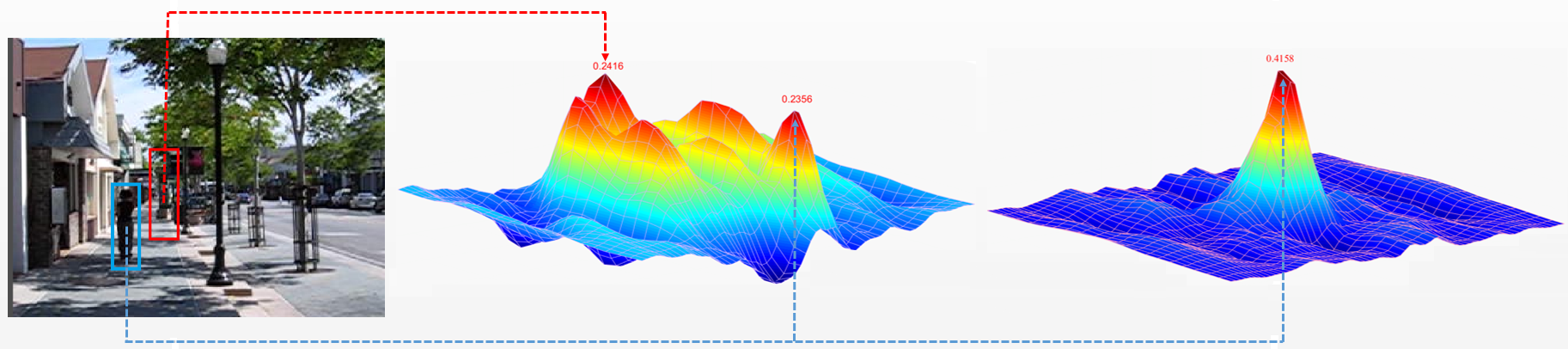}}
\end{center}
   \caption{Illustration of multimodal target detection in sequence \emph{human9} from OTB-15 \cite{wu2015object}. The blue bounding box indicates the correct location of target, the red one is an incorrect detection. The response of the target is weaker than the background area within the red bounding box as shown in the middle. The unimodal detection will regard the highest peak as the target leading to false detection. The proposed multimodal target detection will redetect the areas centered at other peaks to find the maximum peak among these response maps as the right subfigure and locate the correct position of the target.}
\label{LMCF¡ªUni}
\end{figure*}

There are two variables ${\bf{w}}$ and ${\bf{z}}$ to be solved in Eq.\ref{optimization_final}. Whenever one of them is known, the subproblem on the other has a closed form solution. Thus similar to \cite{zuo2016learning}, we introduce the alternating optimization algorithm to solve the model efficiently by iterating between the following two steps.

\noindent\textbf{Update ${\bf{z}}$}. Given ${\bf{w}}$, the subproblem on ${\bf{z}}$ becomes:
\begin{equation}
\mathop {\min }\limits_{\bf{z}} \left\| {{\bf{z}} - \left( {{{\bf{w}}^T}{\Phi _0} - {{\bf{w}}^T}\Phi  - \Upsilon } \right)} \right\|_2^2,{\text{s}}{\text{.t}}{\text{. }}{\bf{z}} \geqslant 0
\end{equation}
Then the closed form solution of ${\bf{z}}$ is:
 \begin{equation}\label{solution_z}
{\bf{z}} = \max \left\{ {{{\bf{w}}^T}{\Phi _0} - {{\bf{w}}^T}\Phi  - \Upsilon ,0} \right\}
\end{equation}

\noindent\textbf{Update ${\bf{w}}$}. Given ${\bf{z}}$, the subproblem on ${\bf{w}}$ becomes:
\begin{equation}
\mathop {\min }\limits_{\bf{w}} \frac{1}{2}{\left\| {\bf{w}} \right\|^2} + C\left\| {{{\bf{w}}^T}\Phi  - \left( {{{\bf{w}}^T}{\Phi _0} - \Upsilon  - {\bf{z}}} \right)} \right\|_2^2
\end{equation}
In order to employ the correlation filter theory, we define ${{\bf{u}}_0} = {{\bf{w}}^T}{\Phi _0}$ which stands for a plane whose height is the highest peak of ${{{\bf{w}}^T}\Phi }$ in the last iteration. Then the closed form solution of ${\bf{w}}$ is:
 \begin{equation}\label{solution_w}
{\bf{\hat w}} = \frac{{{{\hat \Psi }^*}\left( {{\bf{x}},{{\bf{y}}_0}} \right) \circ {{{\bf{\hat u}}}^T}}}{{{{\hat \Psi }^*}\left( {{\bf{x}},{{\bf{y}}_0}} \right) \circ \hat \Psi \left( {{\bf{x}},{{\bf{y}}_0}} \right) + \frac{1}{{2C}}}}
\end{equation}
where ${\bf{u}} = {{\bf{u}}_0} - \Upsilon  - {\bf{z}}$ and $\frac{ \bullet }{ \bullet }$ denotes the element-wise division.

\noindent\textbf{Nonlinear extension}. The proposed linear model can be extended to a nonlinear model by the kernel trick ${K_{ij}} = \left\langle {\varphi \left( {\Psi \left( {{\bf{x}},{{\bf{y}}_i}} \right)} \right),\varphi \left( {\Psi \left( {{\bf{x}},{{\bf{y}}_j}} \right)} \right)} \right\rangle $£¬ where $\varphi \left(  \bullet  \right)$ indicates the implicit use of a high-dimensional feature space. The solution w can be represented as ${\bf{w}} = \sum\limits_{w = 0}^{W - 1} {{\alpha _w}\varphi \left( {\Psi \left( {{\bf{x}},{{\bf{y}}_w}} \right)} \right)} $.
The optimization now is rewritten as
\begin{equation}\label{nonlinear}
\begin{aligned}
 &\mathop {\min }\limits_{\bf{\alpha }} {{\bf{\alpha }}^T}{\mathcal F}^{{\text{ - }}1}\left( {{{{\bf{\hat k}}}^{{\Psi _0}{\Psi _0}}} \circ {\bf{\hat \alpha }}} \right) \\
 &+ C\left\| {{{\mathcal F}^{ - 1}}\left( {{{{\bf{\hat k}}}^{{\Psi _0}{\Psi _0}}} \circ {\bf{\hat \alpha }}} \right) - {{\left( {{{\bf{u}}_0} - \Upsilon  - {\bf{z}}} \right)}^T}} \right\|_2^2\\
&\hspace{1cm}{\text{s}}{\text{.t}}{\text{. }}{\bf{z}} \geqslant 0
\end{aligned}
\end{equation}
where ${\Psi _0} = \Psi \left( {{\bf{x}},{{\bf{y}}_0}} \right)$ and ${{{{\bf{\hat k}}}^{{\Psi _0}{\Psi _0}}}}$ denotes the DFT of the first row of the circulant kernel matrix ${\bf{K}}$ whose elements are ${K_{ij}}$. The closed form of the subproblem on ${\bf{\alpha }}$ is
\begin{equation}\label{solution_a}
{\bf{\hat \alpha }} = \frac{{{{{\bf{\hat u}}}^T}}}{{{{{\bf{\hat k}}}^{{\Psi _0}{\Psi _0}}} + \frac{1}{{2C}}}}
\end{equation}
where $\frac{ \bullet }{ \bullet }$ denotes the element-wise division.
%-------------------------------------------------------------------------
\subsection{Multimodal target detection}
Intuitively, when a new frame comes out, the transformation of the target ${\bf{y}}=f\left( {{\bf{s}};{\bf{w}}} \right)$ is estimated by the Eq.\ref{function_f}, where ${\bf{s}}$ is the region in the new frame centered at the target position of the last frame. This can be sped up with the learned model by FFT algorithm. The full detection response map on all cyclic transform is obtained by
\begin{equation}\label{response_map}
F\left( {{\bf{s}},{\bf{y}};{\bf{w}}} \right) = {{\mathcal F}^{ - 1}}\left( {\hat \Psi _{{\bf{s}}0}^* \circ {\bf{\hat w}}} \right) = {{\mathcal F}^{ - 1}}\left( {{{{\bf{\hat k}}}^{{\Psi _{{\bf{x}}0}}{\Psi _{{\bf{s}}0}}}} \circ {\bf{\hat \alpha }}} \right)
\end{equation}
where ${\Psi _{ \bullet 0}}$ is short for $ \Psi \left( { \bullet ,{{\bf{y}}_{0,0}}} \right)$. The localization of the target is estimated on the highest peak of the response map which is defined as the unimodal detection in this paper. However, the unimodal detection may be disturbed by similar objects or certain noise leading to inaccurate detection. The inaccurate detection would further contaminate the learned model due to incorrect training samples. Shown as Figure \ref{LMCF¡ªUni}, the peaks located at similar objects or background noise in the response map may approach, or even surpass the peak at the target. As above analysis, the target may locate at one of multiple peaks, all of them should be taken into consideration.

Consequently, a multimodal target detection method is proposed to improve localization precision further. For the unimodal detection response map $F\left( {{\bf{s}},{\bf{y}};{\bf{w}}} \right)$, the multiple peaks are computed by
\begin{equation}\label{regional_maxima}
P\left( {\bf{s}} \right) = F\left( {{\bf{s}},{\bf{y}};{\bf{w}}} \right) \circ {\bf{B}}
\end{equation}
where ${\bf{B}}$ is a binary matrix with the same size as $F\left( {{\bf{s}},{\bf{y}};{\bf{w}}} \right)$, which identifies the locations of local maxima in $F\left( {{\bf{s}},{\bf{y}};{\bf{w}}} \right)$. The elements at the locations of local maxima in ${\bf{B}}$ are set to 1, while others are set to 0. All non-zero elements in $P\left( {\bf{s}} \right)$ indicate multiple peaks in the response map of ${\bf{s}}$.

When the ratios between multiple peaks to the highest peak are greater than a predefined threshold $\theta $, the corresponding image regions centered at those peaks are re-detected through Eq.\ref{response_map}. The target is finally identified to locate at the maximum peak among these response maps as shown in Figure \ref{LMCF¡ªUni}.

Furthermore, to handle scale variation, we adopt a scale searching strategy proposed by \cite{danelljan2014accurate} at the detected location. The difference between ours and \cite{danelljan2014accurate} lies in that the scale model is only executed when the detected results have high-confidence as discussed in the next section.
%-------------------------------------------------------------------------
\subsection{High-confidence update}
Most existed trackers update tracking models \cite{danelljan2014accurate,ningobject,henriques2015high,staple_2016_CVPR} at each frame without considering whether the detection is accurate or not. Actually, this may cause a deterministic failure once the target is detected inaccurately, severely occluded or totally missing in the current frame. In the proposed method, we utilize the feedback from tracking results during target detection to decide the necessity of model update.
\begin{figure}[t]
\begin{center}
\subfigure  [Occlusion]{\includegraphics[height=0.8in,width=0.9in,angle=0]{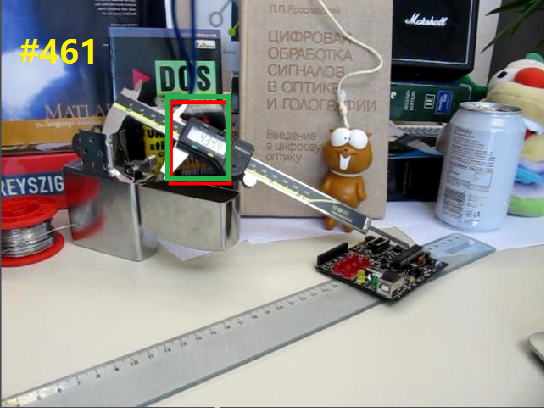}}
\subfigure [No update]{\includegraphics[height=0.8in,width=1.15in,angle=0]{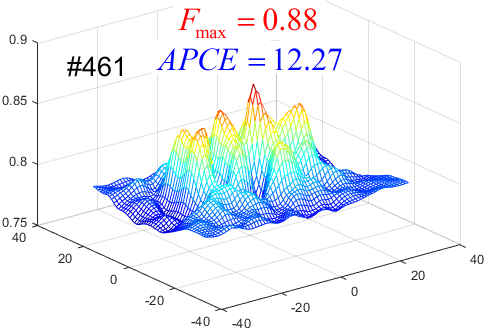}}
\subfigure [Update]{\includegraphics[height=0.8in,width=1.15in,angle=0]{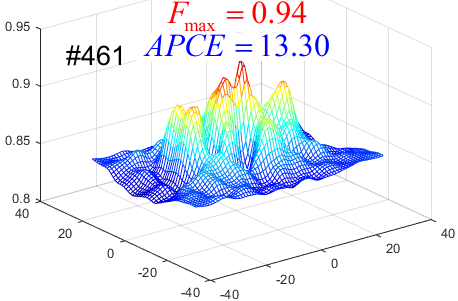}}
\subfigure [No occlusion]{\includegraphics[height=0.8in,width=0.9in,angle=0]{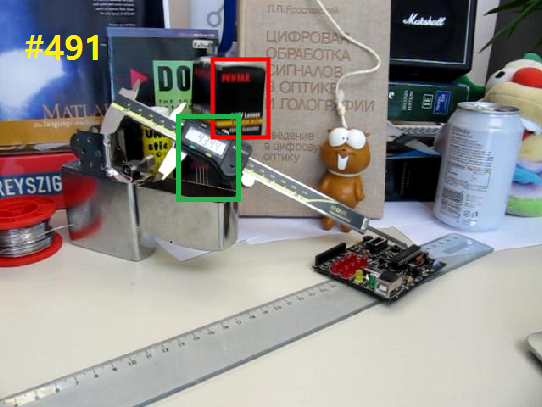}}
\subfigure [Correct tracking]{\includegraphics[height=0.8in,width=1.15in,angle=0]{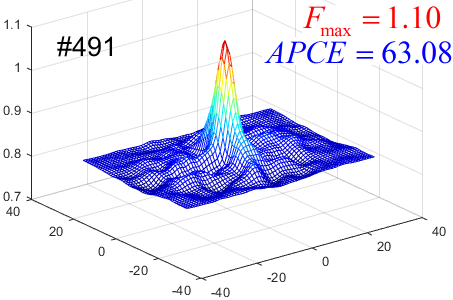}}
\subfigure [Incorrect tracking]{\includegraphics[height=0.8in,width=1.15in,angle=0]{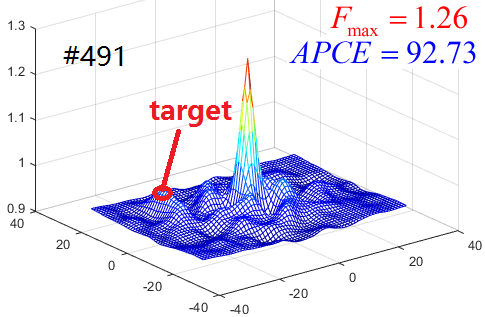}}
%\subfigure {\includegraphics[height=0.8in,width=0.9in,angle=0]{Box_511_LMCF.png}}
%\subfigure {\includegraphics[height=0.8in,width=1.15in,angle=0]{Box_511_map_LMCF.png}}
%\subfigure {\includegraphics[height=0.8in,width=1.15in,angle=0]{Box_511_map_NU.png}}
%\subfigure {\includegraphics[height=0.8in,width=0.9in,angle=0]{Box_595_LMCF.png}}
%\subfigure {\includegraphics[height=0.8in,width=1.15in,angle=0]{Box_595_map_LMCF.png}}
%\subfigure {\includegraphics[height=0.8in,width=1.15in,angle=0]{Box_595_map_NU.png}}
\end{center}
   \caption{The first column are the shots of sequence \emph{box} from OTB-15, where the red bounding boxes indicate the tracking results of LMCF with high-confidence update strategy and the green ones belong to the LMCF-NU which updates the tracking model in each frame. The response maps in the second column are corresponding to LMCF and the third column corresponding to LMCF-NU. The red annotation in the last subfigure points out the right position of the target in this response map.}
\label{LMCF¡ªNU}
\end{figure}

The peak value and the fluctuation of the response map can reveal the confidence degree about the tracking results to some extent. The ideal response map should have only one sharp peak and be smooth in all other areas when the detected target is extremely matched to the correct target. The sharper the correlation peaks are, the better the location accuracy is. Otherwise, the whole response map will fluctuate intensely, whose pattern is significantly different from normal response maps as shown in the first row of Figure \ref{LMCF¡ªNU}. If we continue to use uncertain samples to update the tracking model, it would be corrupted mostly as shown in the second row of the Figure \ref{LMCF¡ªNU}. So we explore a high-confidence feedback mechanism with two criteria. The first one is the maximum response score ${F_{\max }}$ of the response map $F\left( {{\bf{s}},{\bf{y}};{\bf{w}}} \right)$ defined as
\begin{equation}\label{ymax}
{F_{\max }} = \max F\left( {{\bf{s}},{\bf{y}};{\bf{w}}} \right)
\end{equation} 
The second one is a novel criterion called average peak-to-correlation energy (APCE) measure which is defined as
\begin{equation}\label{APCE}
APCE = \frac{{{{\left| {{F_{\max }} - {F_{\min }}} \right|}^2}}}{{mean\left( {\sum\limits_{w,h} {{{\left( {{F_{w,h}} - {F_{\min }}} \right)}^2}} } \right)}}
\end{equation}
where ${F_{\max }}$, ${{F_{\min }}}$ and ${{F_{w,h}}}$ denote the maximum, minimum and the $w$-th row $h$-th column elements of $F\left( {{\bf{s}},{\bf{y}};{\bf{w}}} \right)$. APCE indicates the fluctuated degree of response maps and the confidence level of the detected targets. For sharper peaks and fewer noise, i.e., the target apparently appearing in the detection scope, APCE will become larger and the response map will become smooth except for only one sharp peak. Otherwise, APCE will significantly decrease if the object is occluded or missing.

When these two criteria ${F_{\max }}$ and APCE of the current frame are greater than their respective historical average values with certain ratios ${\beta _1}$, ${\beta _2}$, the tracking result in the current frame is considered to be high-confidence. Then the proposed tracking model will be updated online with a learning rate parameter $\eta$ as
\begin{equation}\label{update}
\begin{aligned}
 &{{{\bf{\hat \alpha }}}^t} = \left( {1 - \eta } \right){{{\bf{\hat \alpha }}}^{t - 1}} + \eta {\bf{\hat \alpha }} \\
 &{{\hat \Psi }^t}_{{\bf{x}}0} = \left( {1 - \eta } \right)\hat \Psi _{{\bf{x}}0}^{t - 1} + \eta {{\hat \Psi }_{{\bf{x}}0}}
\end{aligned}
\end{equation}

Figure \ref{LMCF¡ªNU} illustrates the importance of the proposed update strategy. As shown in Figure \ref{LMCF¡ªNU}, when the target is occluded severely, the response map fluctuates fiercely in the first row so that APCE reduces to about 10, while ${F_{\max }}$ remains strong enough. Under this circumstance, the proposed high-confidence update strategy will choose not to update the model in this frame, then the tracking model won't be corrupted and the target can be tracked successfully in the subsequent frames. Otherwise, the target will be missed and the right peak will finally fade away.

An overview of the proposed method is summarized in Algorithm \ref{algorithm}.

\begin{algorithm}
\caption{LMCF tracking algorithm} \label{algorithm}
\small{
\begin{algorithmic}[1]
    \REQUIRE{Frames ${\left\{ {{{\bf{I}}_t}} \right\}_1^T}$, initial target location ${{\bf{p}}_1}$, ${\bf{z}} = 0$, ${{\bf{u}}_0} = ones\left( {W,H} \right)$}
    \ENSURE{Target locations of each frame $\left\{ {{{\bf{p}}_t}} \right\}_2^T$.}
\REPEAT
\STATE Crop an image region ${\bf{s}}$ from ${{\bf{I}}_t}$ at the last location ${{\bf{p}}_{t - 1}}$ and extract its joint feature map $\Psi \left( {{\bf{s}},{{\bf{y}}_{0,0}}} \right)$.
\STATE Detect the target location ${{{\bf{p}}_t}}$ with the multimodal detection via Eq.\ref{response_map} and Eq.\ref{regional_maxima}.
\STATE Estimate the scale of the target as \cite{danelljan2014accurate}.
\STATE Calculate ${F_{\max }}$ and APCE with Eq.\ref{ymax} and Eq.\ref{APCE}.
   \IF {${F_{\max }}$ and APCE satisfy the update condition,} 
        \STATE Train the ${{\bf{u}}_0}$, ${\bf{z}}$ and ${{\bf{\hat w}}}$ $\left( {{\bf{\hat \alpha }}} \right)$ with Eq.\ref{solution_z} and Eq.\ref{solution_w} (\ref{solution_a}).
        \STATE Update the tracking model with Eq.\ref{update}.
        \STATE Update the scale estimation model as \cite{danelljan2014accurate} with $\eta $.
    \ENDIF
\UNTIL end of video sequence.
\end{algorithmic}
}
\end{algorithm}

%-------------------------------------------------------------------------
\section{Experiments}
Since the proposed tracking algorithm is compatible with different kinds of features for representing the targets, we implement experiments with both conventional features based version LMCF and deep CNNs based version DeepLMCF to validate the performance of the proposed method.

We implement experiment on the OTB-13 \cite{wu2013online} and OTB-15 \cite{wu2015object} benchmark datasets. All these sequences are annotated with 11 attributes which cover various challenging factors, including scale variation (SV), occlusion (OCC), illumination variation (IV), motion blur (MB), deformation (DEF), fast motion (FM),  out-of plane rotation (OPR), background clutters (BC), out-of-view (OV), in-plane rotation (IPR) and low resolution (LR). To fully assess our method, we use one-pass evaluation (OPE), temporal robustness evaluation (TRE), and spatial robustness evaluation (SRE) metrics as suggested in \cite{wu2013online}. The \emph{precision} scores indicate the percentage of frames in which the estimated locations are within 20 pixels compared to the ground-truth positions. The \emph{success} scores are defined as the area
under curve (AUC) of each success plot, which is the average of the success rates corresponding to the sampled overlap threshold.

We first analyze LMCF with the improvements from multimodal target detection, high-confidence update strategy and representation power of DeepLMCF on OTB-13. Then we compare LMCF with 9 most related and state-of-the-art trackers based on conventional features on OTB-13 and OTB-15. Finally, we present the attractive performance of DeepLMCF compared with 9 up-to-date CNNs based trackers on OTB-13. All the tracking results are using the reported results to ensure a fair comparison.

\subsection{Implementation details}
 The conventional features used for LMCF are composed of HOG features and color names (CN) \cite{danelljan2014adaptive}. For the CNN features of DeepLMCF, we use imagenet-vgg-verydeep-19 which is available at: http://www.vlfeat.org/matconvnet/. The last three convolutional layers of this network are used to extract the features of the target and the weight of each layer is respectively set to 0.02, 0.5 and 1 similar to \cite{HCF2015hierarchical}. Our tracker is implemented in MATLAB for LMCF with a PC with a 3.60 GHz CPU and DeepLMCF with a tesla k40 GPU. LMCF runs faster than 80 FPS while DeepLMCF runs faster than 10 FPS.

The optimization takes 10 iterations in the first frame and 3 iterations for each online update. Similar to \cite{danelljan2014accurate}, 33 number of scales with a scale factor of 1.02 is used in the scale model. The other parameters setting of LMCF and DeepLMCF are shown in Table \ref{tab:parameters}, where padding means the magnification of the image region samples relative to the target bounding box.
%\begin{table}
%\caption{\label{tab:parameters}Parameters setting of LMCF and DeepLMCF.}
%	\begin{center}
%		\begin{tabular}{|c|c|c|c|c|c|c|}
%			\hline
%			    &$\eta $&$\theta $&${\beta _1}$& ${\beta _2}$&padding&C \\
%			\hline
%			LMCF & 0.015&0.7&0.65&0.55&1.5&10000 \\
%			\hline
%			DeepLMCF &0.01&0.7&0.4&0.3&1.8&10000 \\
%			\hline
%		\end{tabular}
%	\end{center}
%\end{table}
\begin{table}
\caption{\label{tab:parameters}Parameters of LMCF and DeepLMCF.}
	\begin{center}
		\begin{tabular}{c c c}
			\hline
                parameters &LMCF & DeepLMCF \\
            \hline
                padding &1.5 &1.8\\
                $\eta $ & 0.015& 0.01\\
                $\theta $& 0.7&0.7\\
                ${\beta _1}$& 0.7 &0.4\\               
			    ${\beta _2}$& 0.45&0.3\\
                C &10000 &20000\\
			\hline
		\end{tabular}
	\end{center}
\end{table}

%-------------------------------------------------------------------------
\subsection{Analyses of LMCF}
To demonstrate the effect of the proposed multimodal target detection, high-confidence update strategy and representation power of DeepLMCF, we first test with different versions of LMCF on OTB-13. We denote LMCF without multimodal detection as LMCF-Uni, without high-confidence update strategy as LMCF-NU and with neither of these two as LMCF-N2. The characteristics and tracking results are summarized in Table \ref{tab:Analysis}. The mean FPS here is estimated on the longest sequence \emph{doll} in OTB-13 with 3872 frames.

As shown in Table \ref{tab:Analysis}, DeepLMCF shows the best tracking accuracy and robustness in all OPE, TRE and SRE evaluation metrics benefited by the hierarchical CNN features and LMCF performs second while with the fastest speed. Without multimodal detection, LMCF-Uni gets poor performance because of false detection from similar objects or background noise. Additionally, incorrect results are likely leading to unwanted updates, resulting in the fact that operating efficiency is lower than LMCF. Without high-confidence update strategy, LMCF-NU updates the tracking model in each frame, thus the tracking speed is dramatically reduced to nearly half to LMCF and the accuracy is also less than LMCF. Without both of these two, LMCF-N2 reaches the last one in all evaluation metrics. Although the proposed multimodal detection increases the detection time, our high-confidence update strategy speeds up the model update process significantly. Both of them improve the tracking performance observably according to the experimental results.

\begin{table*}\footnotesize
	\caption{\label{tab:Analysis}Characteristics and tracking results of LMCF, DeepLMCF, LMCF-Uni, LMCF-NU and LMCF-N2. The entries in {\color{red}red} denote the best results and the ones in {\color{blue}blue} indicate the second best.}
	\begin{center}
		\begin{tabular}{|c|c|c|c|c|c|c|c|c|c|c|}
	\hline
	\multirow{2}{*}{Trackers }& multimodal  & high-confidence& feature& \multicolumn{2}{|c|}{OPE} & \multicolumn{2}{|c|}{TRE} & \multicolumn{2}{|c|}{SRE} & mean \\
\cline{5-10}
	&detection& update& representations&precision&success&precision&success&precision&success&FPS\\
   \hline
    LMCF-N2& No&No& conventional&0.799&0.586&0.813&0.612&0.740&0.540&60.74\\
   \hline
	LMCF-Uni& No&Yes& conventional&0.809&0.606&0.815&0.616&0.757&0.549&{\color{blue}61.38}\\
   \hline
    LMCF-NU& Yes&No& conventional&0.813&0.605&0.820&0.619&0.750&0.545&46.45\\
   \hline
   LMCF & Yes&Yes& conventional&{\color{blue}0.839}&{\color{blue}0.624}&{\color{blue}0.829}&{\color{blue}0.625}&{\color{blue}0.760}&{\color{blue}0.552}&{\color{red}85.23}\\
    \hline
   DeepLMCF&Yes&Yes& deep CNNs&{\color{red}0.892}&{\color{red}0.643}&{\color{red}0.877}&{\color{red}0.649}&{\color{red}0.850}&{\color{red} 0.596}&8.11\\
    \hline
		\end{tabular}
	\end{center}
\end{table*}

\subsection{Evaluation on LMCF}
We evaluate LMCF with 9 state-of-the-art trackers designed with conventional hand-crafted features including Struck \cite{hare2011struck}, MEEM \cite{zhang2014meem}, TGPR \cite{TGPR2014transfer}, DLSSVM \cite{ningobject}, Staple \cite{staple_2016_CVPR}, KCF \cite{henriques2015high}, RPT \cite{RPT_2015_CVPR}, DSST \cite{danelljan2014accurate} and SAMF \cite{SAMF2014scale}. Among them, Struck and DLSSVM are structured SVM based methods, Staple, KCF, DSST, RPT and SAMF are CF based algorithms, MEEM and TGPR are developed based on regression
and multiple trackers.
\begin{figure}[t]
\begin{center}
   \subfigure {\includegraphics[height=1.2in,width=1.6in,angle=0]{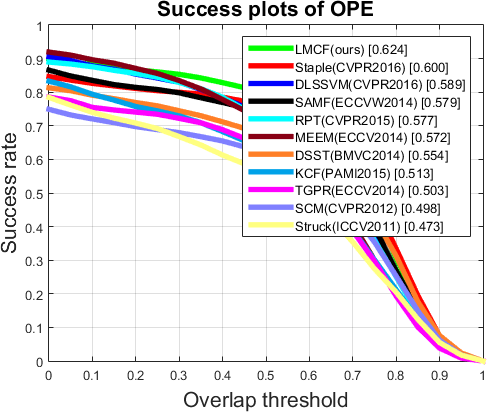}}
\subfigure {\includegraphics[height=1.2in,width=1.6in,angle=0]{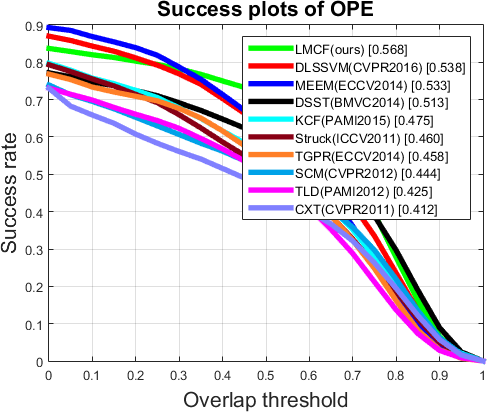}}
\subfigure {\includegraphics[height=1.2in,width=1.6in,angle=0]{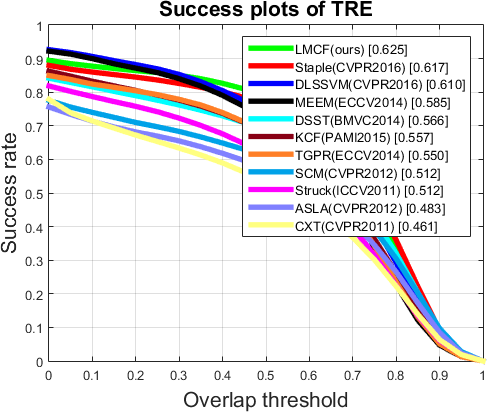}}
\subfigure {\includegraphics[height=1.2in,width=1.6in,angle=0]{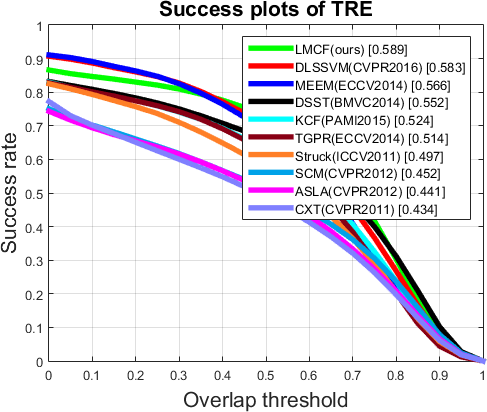}}
\subfigure {\includegraphics[height=1.2in,width=1.6in,angle=0]{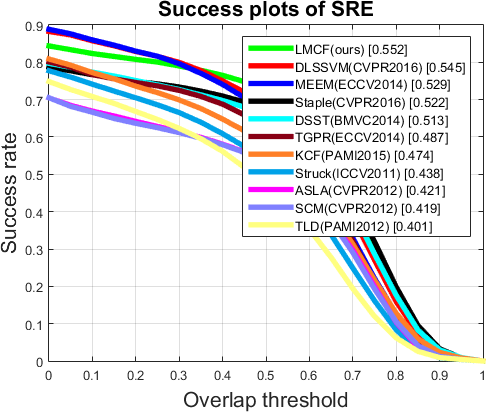}}
\subfigure {\includegraphics[height=1.2in,width=1.6in,angle=0]{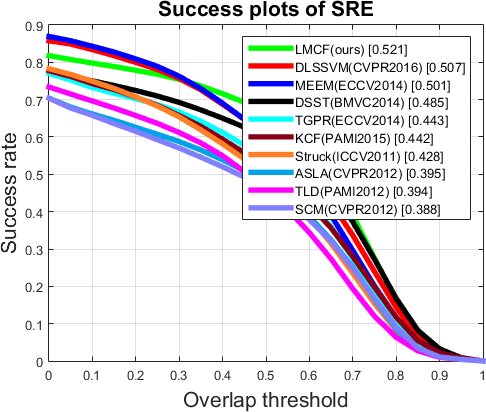}}
\end{center}
   \caption{The success plots of OPE, TRE, SRE on OTB-13 (left column) and OTB-15 (right column). The numbers in the legend indicate the average AUC scores for success plots. The years and original sources of these trackers are also shown in the legend. Results are best viewed on high-resolution displays.}
\label{LMCF_evaluation}
\end{figure}

Figure \ref{LMCF_evaluation} illustrates the success plots of top ten trackers on both OTB-13 and OTB-15. LMCF performs best with all OPE, TRE and SRE evaluation metrics in the two benchmarks. Struck performed the first when the original benchmark \cite{wu2013online} first came out, so that it is a good representation of its previous trackers. LMCF significantly improves Struck by an average improvement of 15\% in the average AUC scores. The DSST and SAMF mainly focus on the scale estimation, their speed are 24 FPS and 7 FPS as they reported. Our method employs the scale estimation method from DSST, but the proposed LMCF performs favorably over the DSST as well as SAMF while runs more than 3 times faster than DSST and more than 11 times faster than SAMF. As for tracking efficiency, Staple and KCF are the only two with comparable reported speeds of 80 FPS and 172 FPS, while LMCF outperforms them in all evaluations. Moreover, LMCF is also superior to other up-to-date trackers like MEEM, TGPR, RPT, SAMF and DLSSVM with a significantly higher speed.

\begin{figure*}[t]
\begin{center}
\subfigure {\includegraphics[height=1.2in,width=1.65in,angle=0]{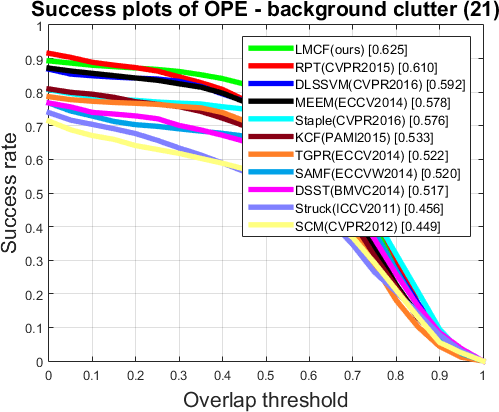}}
\subfigure {\includegraphics[height=1.2in,width=1.65in,angle=0]{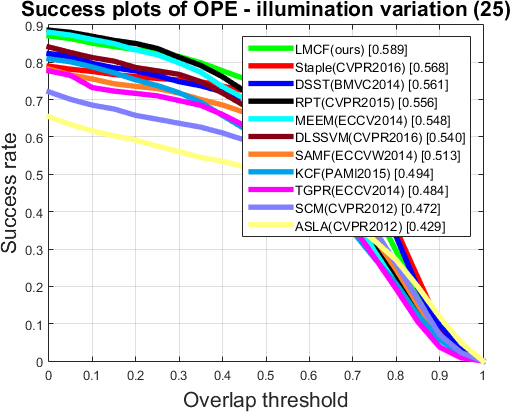}}
\subfigure {\includegraphics[height=1.2in,width=1.65in,angle=0]{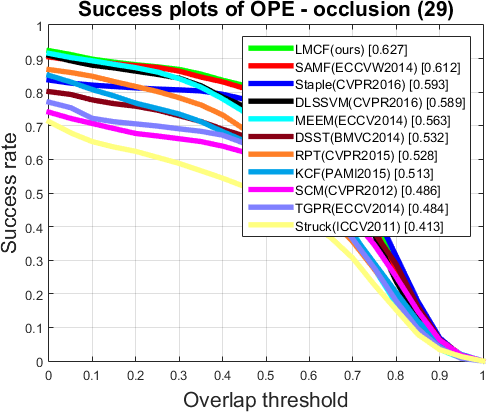}}
\subfigure {\includegraphics[height=1.2in,width=1.65in,angle=0]{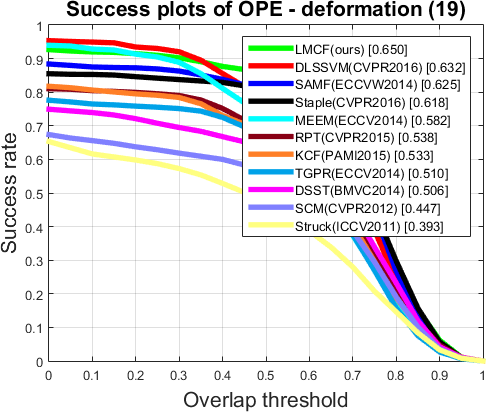}}
\subfigure {\includegraphics[height=1.2in,width=1.65in,angle=0]{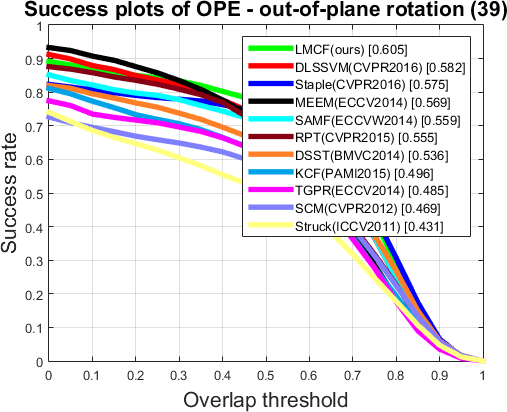}}
\subfigure {\includegraphics[height=1.2in,width=1.65in,angle=0]{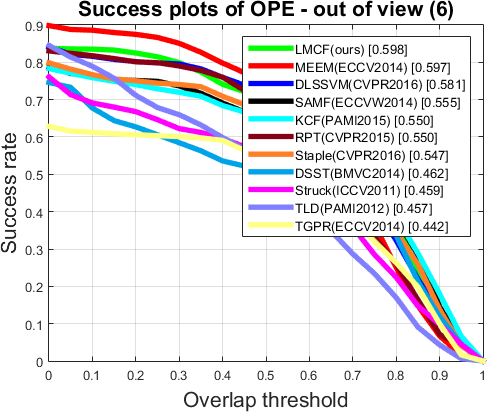}}
\subfigure {\includegraphics[height=1.2in,width=1.65in,angle=0]{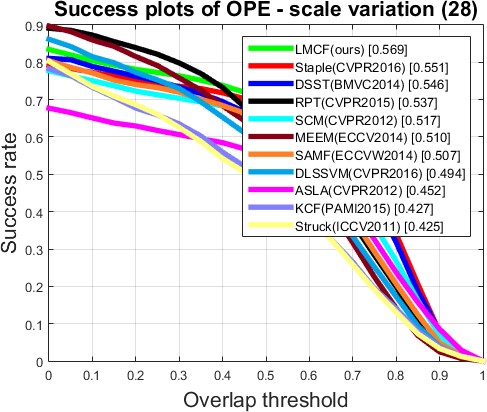}}
\subfigure {\includegraphics[height=1.2in,width=1.65in,angle=0]{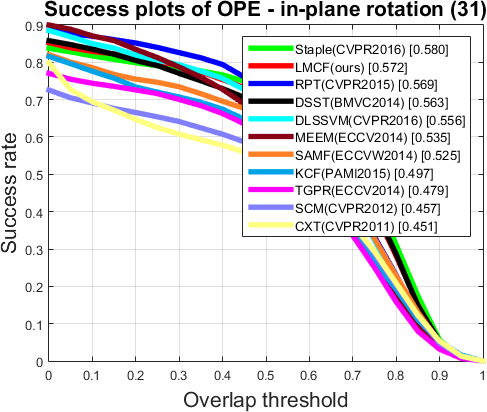}}
\end{center}
    \caption{The success plots for 8 challenging attributes including background clutter, illumination variation, occlusion, deformation, out-of-plane rotation, out-of-view, scale variation and in-plane rotation. The proposed LMCF performs best in almost all the attributes. Results are best viewed on high-resolution displays.}
\label{att}
\end{figure*}

For detailed analyses, we also evaluate LMCF with these trackers on various challenging attributes in OTB-13 as shown in Figure \ref{att}. The results demonstrate that LMCF performs well on most attributes, especially on occlusion, scale variation, illumination variation, background clutter and out of plane rotation.

\subsection{Evaluation on DeepLMCF}

\begin{figure}[t]
\begin{center}
    {\includegraphics[height=1.2in,width=1.6in,angle=0]{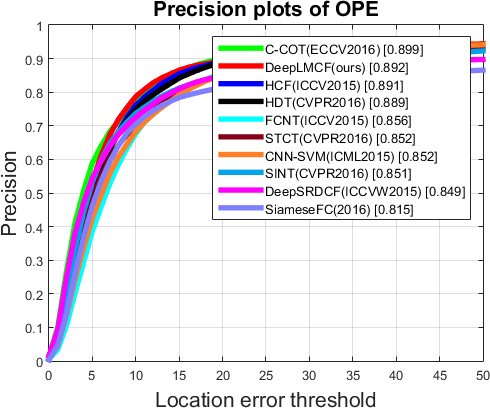}}
 {\includegraphics[height=1.2in,width=1.6in,angle=0]{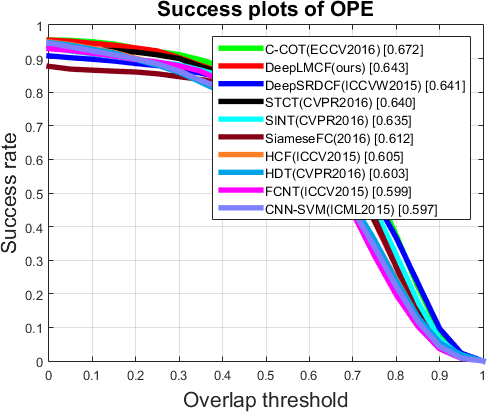}}
\end{center}
   \caption{The precision and success plot of OPE on OTB-13. The numbers in the legend indicate the average precision scores for precision plot and the average AUC scores for success plot. Results are best viewed on high-resolution displays.}
\label{CNN}
\end{figure}
To further improve the tracking accuracy and robustness of LMCF, we implement DeepLMCF with deep CNNs based features. It is compared with 9 up-to-date CNNs based trackers including C-COT \cite{C-COT2016beyond}, DeepSRDCF\cite{DeepSRDCF2015convolutional}, HCF\cite{HCF2015hierarchical}, HDT\cite{HDT2016hedged}, STCT\cite{STCT2016stct}, CNN-SVM\cite{icml2015_hong15}, SINT\cite{tao2016sint}, FCNT\cite{FCNT2015visual} and SiameseFC\cite{SiameseFC2016fully}.

Figure \ref{CNN} demonstrates the performance of DeepLMCF with the 9 CNNs based trackers on OTB-13. Although the proposed DeepLMCF scores the second following the C-COT tracker on the precision and success scores, the tracking speed of DeepLMCF is 40 times faster than C-COT with the speed from its reported results at about 0.25 FPS, which is a severe limitation of its application. The most related method to DeepLMCF is HCF due to the similar feature hierarchy. But DeepLMCF keeps ahead of it especially on success score mainly because the scale variations of the target are not considered by HCF. Moreover, HCF and SiameseFC are the only two with comparable reported speeds of 10 FPS and 58 FPS, while LMCF performs superiorly against them in both evaluations. In summary, the proposed DeepLMCF outperforms these trackers except for C-COT while remains a comparably fast speed at more than 10 FPS.

%\begin{table}
%\begin{center}
%\caption{Results.   Ours is better.}
%\begin{tabular}{|c| c|c |c|}
%\hline
%\multicolumn{2}{|c|}{Conventional Features} & \multicolumn{2}{|c|}{CNNs Features} \\
%\hline
%Trackers & Speed & Trackers & Speed \\
%\hline
%Struck & Frumpy& CNN-SVM & Frumpy \\
%MEEM & Frumpy& HDT & Frumpy \\
%DSST & Frumpy& STCT & Frumpy \\
%SAMF & Frumpy& SINT & Frumpy \\
%RPT & Frumpy& FCNT & Frumpy \\
%TGPR & 3~4& C-COT & <1 \\
%Staple & 80& SiameseFC & Frumpy \\
%KCF & 172& HCF & Frumpy \\
%SDLSSVM & 5.40& DeepSRDCF & Frumpy \\
%LMCF & 60 & DeepLMCF & Frumpy \\
%\hline
%\end{tabular}
%\end{center}
%\end{table}

%------------------------------------------------------------------------
\section{Conclusion}
In this paper, we propose a novel large margin object tracking method with circulant feature maps. A bridge is built up to link the framework with correlation filter. Hence, the proposed LMCF tracker absorbs the strong discriminative ability from structured output SVM and speeds up by the correlation filter algorithm significantly. In order to prevent model drift introduced by similar objects or background noise, a multimodal target detection technique is proposed to ensure the correct detection. Moreover, we establish a high-confidence model update strategy to avoid the model corruption problem. Furthermore, the proposed tracking algorithm is equipped with strong compatibility, thus we also implement a deep CNNs based version DeepLMCF to verify its outstanding performance. Sufficient evaluations on challenging benchmark datasets demonstrate that the proposed LMCF and DeepLMCF tracking algorithms perform well against most state-of-the-art methods including both conventional features and deep CNNs features based trackers. It is worth to emphasize that our proposed algorithm not only performs superiorly, but also runs at a very fast speed which is sufficient for realtime applications.
{\small
\bibliographystyle{ieee}
\bibliography{egbib}
}

\end{document}